\documentclass[10pt,twocolumn,a4paper]{article}
\usepackage[T1]{fontenc}

\usepackage{cvpr}
\usepackage{times}
\usepackage{epsfig}
\usepackage{graphicx}
\usepackage{subcaption}

\usepackage{amsmath,amssymb,amsfonts}
\usepackage{bm} 
\usepackage{latexsym}
\usepackage{siunitx}

\usepackage{newunicodechar}
\newunicodechar{◌}{\includegraphics{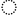}}

\usepackage{placeins}
\usepackage{stfloats}

\usepackage[square,numbers]{natbib}

\usepackage{authblk}

\usepackage{xurl} 
\usepackage[breaklinks=true,bookmarks=false]{hyperref}

\usepackage[capitalize]{cleveref}

\cvprfinalcopy 

\ifcvprfinal\fi

\begin{document}

\title{Blazer: Laser Scanning Simulation using Physically Based Rendering}

\author[1]{Sebastian Grans\thanks{sebastian.grans@ntnu.no}}
\author[1]{Lars Tingelstad}
\affil[1]{Department of Mechanical and Industrial Engineering\protect\\
Norwegian University of Science and Technology}

\maketitle

\begin{abstract}
Line laser scanners are a sub-type of structured light 3D scanners that are relatively common devices to find within the industrial setting, typically in the context of assembly, process control, and welding. Despite its extensive use, scanning of some materials remain a difficult or even impossible task without additional pre-processing. For instance, materials which are shiny, or transparent. 

In this paper, we present a Blazer, a virtual line laser scanner that, combined with physically based rendering, produces synthetic data with a realistic light-matter interaction, and hence realistic appearance. This makes it eligible for the use as a tool in the development of novel algorithms, and in particular as a source of synthetic data for training of machine learning models. Similar systems exist for synthetic RGB-D data generation, but to our knowledge this the first publicly available implementation for synthetic line laser data. 
We release this implementation under an open-source license to aid further research on line laser scanners. 
\end{abstract}

\section{Introduction}

A laser scanner is a type of structured light 3D scanner consisting of a camera in conjunction with a line laser. Through triangulation, the depth at the point where the laser intersects objects in the scene can be deduced. Laser scanners have many applications, especially in industry, but also in cultural heritage preservation. Within industry, applications range from part specification validation, assembly, general automation, and welding to name a few\cite{micro-epsilon}. 

\begin{figure}[!ht]
    \centering
    \includegraphics[width=0.9\linewidth]{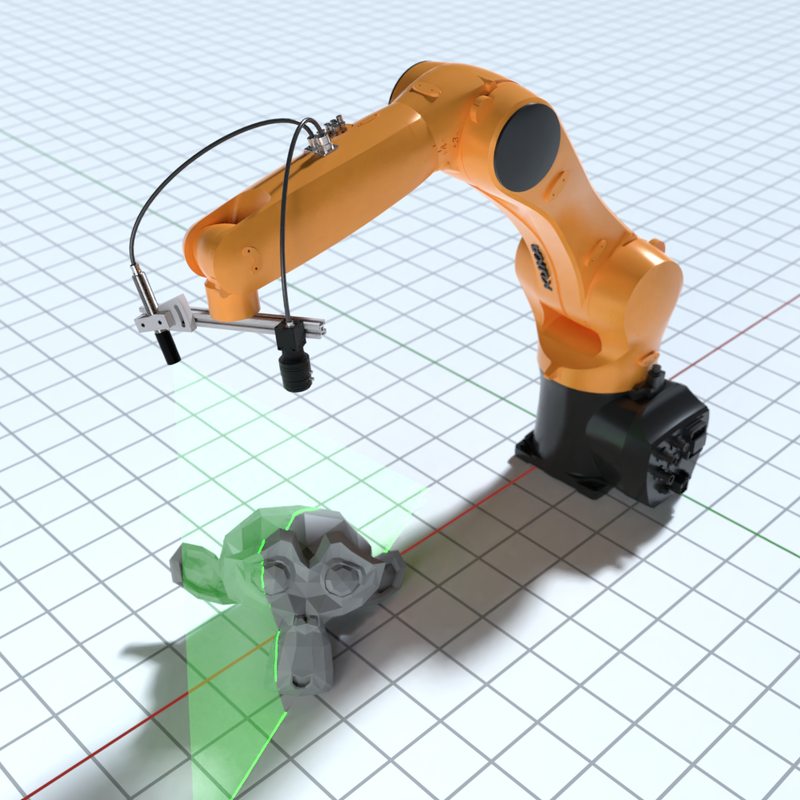}
    \caption{An example of how the virtual laser scanner setup can be used. Here mounted to a fully rigged industrial robot. The green laser plane is purely for visualization purposes.}
    \label{fig:full-setup}
\end{figure}

Within cultural heritage preservation, one of the more famous cases is The Digital Michelangelo Project~\cite{michelangelo}, wherein they used laser 3D scanners to digitize large marble statues. The Smithsonian Institute is another case worthy of mention, as they currently have more than 1700 historic artifacts digitized~\cite{smith}, with laser scanning being one of their 3D reconstruction methods~\cite{faro}. 

Despite the relatively long history of laser scanners being actively used, there are still some materials that are very difficult to correctly 3D reconstruct. These materials are typically reflective, transparent, or produce a lot of subsurface scattering~\cite{reflective-wang, difficult-surfaces, subsurface}. Even though the laser line position might be completely obvious to a human observer, algorithms struggle to isolate it correctly, leading to biased reconstructions at best. In practice, difficult surfaces are often sprayed with anti-reflective coating\cite{reflective-spray}, but this might not always be an option. For instance, when historic artifacts are being scanned, or when the spray might interfere with subsequent processes, such as welding. Hence, further research on improved laser line extraction methods is needed, and this system is designed to aid in that process.

Physically based rendering techniques combines ray-tracing methods, with physics based light-interaction models to produce photorealistic results. These techniques have previously been limited to large scale video productions due to the large computational cost. In recent years, however, graphics hardware has gotten better and cheaper, and now come with hardware-based ray tracing acceleration. This has made physically based rendering techniques (PBR) to be available even on consumer-level hardware. In addition to this, there are free and open-source 3D graphics software such as Blender\textsuperscript{\textregistered}~\cite{blender}, which comes with paired with Cycles, a PBR engine. Hence, enabling anyone to produce photorealistic 3D artwork and simulations. 
In this paper, we leverage the capabilities of Blender to create a virtual laser scanner that can produce simulated scans with high realism. One application of these simulated scans is for training of neural networks. 

Producing real datasets for machine learning applications consumes a lot of time and human resources. Since humans are doing the labeling, the accuracy is also limited. One of the benefits of synthetic data is that it requires less human intervention to produce, and comes with ground truth information by default, free of human bias. 

Synthetic images from the system can also be very useful when developing traditional algorithms. In the context of software development, it is common to have unit testing as part of the development cycle, which ensures that new code does not break previous functionality. Using the type of system presented in this paper, it is very easy to make a base-line test set that the algorithm should always be able to process correctly. As development goes on, one can also easily add new cases with entirely new materials without having to acquire a real sample. Furthermore, synthetic data makes it possible to quantitatively measure the accuracy of an implementation which is typically very difficult to achieve on a physical system.

In both of these cases, it is key that the synthetic data is a good representation of the real data. This is in the field known as the reality gap. 

The main contribution of this paper is the development of an open-source and physically accurate virtual laser scanner implemented in Blender. In \cref{fig:full-setup}, an example of how the virtual setup can be used is shown. The system has the potential to be used as a testing tool for new line laser algorithms, or for validation of existing ones. Another key application is synthetic data generation for training of machine learning models. Similar systems exist for synthetic RGB-D data generation, but to our knowledge, this is the first publicly released implementation designed for the development of line laser scanners.

A note on the notation is used in this paper: 
Bold upright symbols refer to matrices. Bold italic symbols refer to vectors, except for the zero vector, $\bm{0}$, which is upright.

When matrices are written with indices, such as $\mathbf{M}_{ij}$, then the indices indicate between which coordinate frames it transforms. For instance, the product
$$
    \bm{p_i} = \mathbf{M}_{ij}\bm{p}_j
$$
describes the change of basis of a point $\bm{p}$ from the coordinate frame $\{j\}$ to $\{i\}$. Common frames in this paper are $\{w\}$, $\{c\}$, and $\{i\}$, which represent the world, camera, and image frames respectively. Vectors and scalars are sometimes also explicitly written with such an index to indicate in which frame it is represented. 

The structure of the paper is as follows: In \cref{sec:related_works} we give an overview of previous work that has been done in the field of synthetic data generation. We then proceed with \cref{sec:preliminaries} wherein we give a brief introduction to what physically based rendering is and the maths related to laser vision systems. A description of how the virtual laser scanner was constructed is presented in \cref{sec:implementation}, followed by \cref{sec:method} where the details related to how various experiments were performed are contained. The results of these investigations are then presented and discussed in \cref{sec:results_and_disc}, followed by the conclusions in \cref{sec:conclusion}. Finally, in \cref{sec:future} we present some of the outlooks for the system presented in this paper.

The key components of this work, such as relevant source code and the Blender file containing the virtual laser, are released under the MIT license and are available at {\useOriginalUrlSetting \url{https://github.com/SebastianGrans/Blazer}}.

\section{Related work} \label{sec:related_works}
To our knowledge, the closest related work is that of \citet{abu2018virtual}, which implemented a similar system in the 3D graphics software Autodesk 3ds Max\textsuperscript{\textregistered}. They evaluated the system, and the related algorithms, by comparing the virtual one with a real scanner setup. The main difference between our work and theirs is that we focus on realism through PBR to achieve as small of a domain gap as possible, for use in neural network training. We also highlight how our system can be used for developing novel algorithms for difficult materials, in particular reflective ones. In contrast, our system is implemented in the free and open-source 3D software Blender and made publicly available under the MIT license. In \cite{pbr-structured-light}, they implemented a projector-based structured light system in a PBR engine. Using the ground truth data given by the simulation, they could perform quantitative evaluations on various encoding schemes. 

\subsection{Synthetic data}
One of the main purposes of our system is synthetic data generation for machine learning applications. The use of synthetic data is not a new concept in the field as it is a method to cheaply create and augment a dataset. The earliest approaches were based on rendering a 3D object on top of a random background image~\cite{3donimg}. This simplistic approach did however not transfer well to real images, as the reality gap was too large. Later works have been done where entire scenes rendered in 3D were used, which have achieved better result\cite{full-3d-pbr-rendering, domain-rand}. 

The concept of using simulations for training neural networks to then be applied to real data is known as `sim2real'. The concept and its validity for use in robotic applications was heavily discussed at the 2020 Robotics: Science and Systems conference~\cite{sim2realperspectives}.

There currently exist two Blender-based pipelines for synthetic data, namely BlenderProc~\cite{blenderproc} and BlendTorch~\cite{blendtorch}. The former is an offline renderer that focuses on realism through physically based rendering and was used for generating the datasets that were used in the 2020 edition of the BOP challenge~\cite{bop}
. BlendTorch on the other hand is based on real-time rendering and is designed to be used directly inside a PyTorch data loader.

\section{Preliminaries} \label{sec:preliminaries}
\subsection{Physically based rendering}

In this section, a very brief introduction to physically based rendering (PBR) is given. For a full description, please refer to~\cite{pbrt}.

Physically based rendering is the concept of creating an image by simulating how light interacts with the virtual 3D scene by using light-matter interaction models. As will be evident later in the text, this is a very time-consuming process since a large amount of light-rays needs to be simulated. This is contrasted with the rasterization-based rendering techniques in real-time render engines, which are used in video games. These types of render engines use approximations of light transportation and are typically not that good at replicating global illumination which is important for physically accurate renditions of a scene.

In path-tracing algorithms, such as those used by both Cycles~\cite{blender} and LuxCoreRender~\cite{luxcorerender}, light rays are sent out in reverse, i.e., from the camera onto the scene. As the ray hits an object, it will scatter off in a new random direction with a probability described by the materials bidirectional scattering distribution function (BSDF). The ray might then continue to collide with another object, and so on. After a set number of interactions, or if the ray hits a light source, the tracing is terminated. The value of the originating pixel is then a weighted sum of all the materials and lights the ray interacted with. Since the scattering direction is random, each pixel must be sampled multiple times ($>10^2$) to create a proper image, otherwise the resulting image will be very noisy and unrepresentative.

\begin{figure}
    \centering
    \includegraphics[width=.7\linewidth]{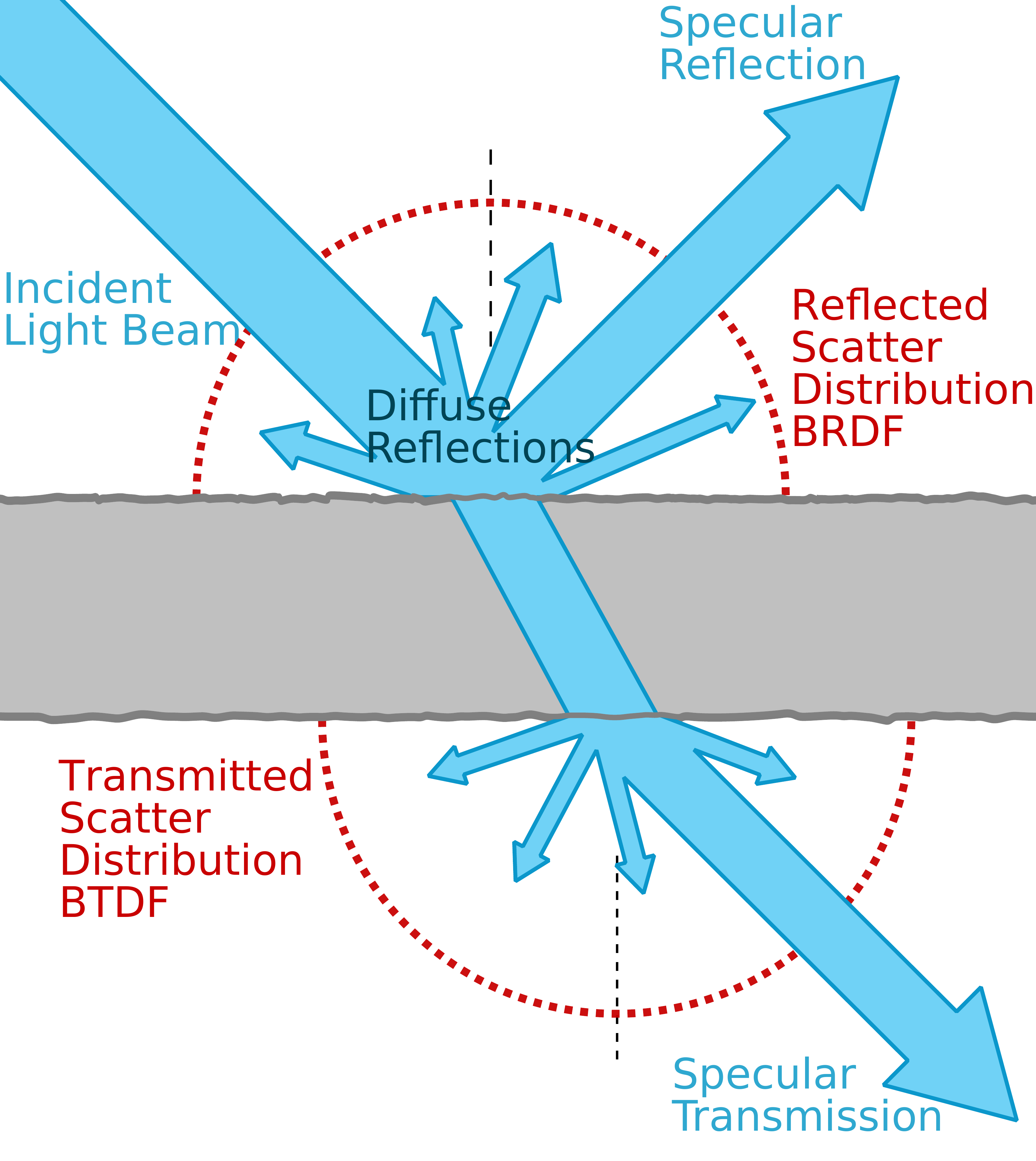}
    \caption{A simplified visualization of the various components of a BSDF. Rework of illustration by Jurohi (License: CC BY-SA 3.0)\protect\footnotemark.}
    \label{fig:prel:bsdf}
\end{figure}

The key component to why this results in a physically accurate depiction is due to the BSDF. The BSDF is a general term distribution functions that describe scattering phenomena, such as reflection, transmission, and subsurface scattering. An illustration of which is shown in \cref{fig:prel:bsdf}.
These functions must have the following properties so as to be physically representative:
\begin{itemize}
    \item Reciprocity: The probability of scattering into a certain angle is identical to scattering in the reverse direction.
    \item Energy conserving: Reflected light must have less or equal amount of energy as the incoming light.  
\end{itemize}
The parameters of the BSDF are most commonly specified by eye until the virtual material looks similar to the real one. This requires a lot of knowledge and experience in order to achieve a realistic result. There are however methods to measure the BSDF of a material~\cite{muller2005acquisition}, and commercial tools and services that perform this also exist~\cite{lighttec}. Hence, partially removing the human from the equation.
\footnotetext{\UrlNoBreaks \url{https://creativecommons.org/licenses/by-sa/3.0/}}

\subsection{Camera model}

The pinhole camera model is the most commonly used model and is a central projection model (c.f. parallel projection). It assumes that only light that travels from the observed scene and through the camera's origin will reach the image plane. It is a mathematical representation of an ideal camera obscura. Mathematically it can be written as 
\begin{equation}
    \bm{p}_i = \mathbf{P} \bm{p}_w \, , \label{eqn:camera_proj}
\end{equation}
where $\mathbf{P}$ is a $3\times4$ projection matrix describing the mapping $\mathbf{P}: \mathbb{P}^3 \mapsto \mathbb{P}^2$. 

The projection matrix $\mathbf{P}$ can be further decomposed into three components
\begin{equation}
    \mathbf{P} = \mathbf{K}[\,\mathbf{I}\,|\,\bm{0}\,]\mathbf{T}_{cw} \, , \label{eqn:camera_proj_full}
\end{equation}
where $\mathbf{K}$ is the intrinsic camera matrix, $\mathbf{T}_{cw}$ the extrinsic camera matrix, and  $[\,\mathbf{I}\,|\,\bm{0}\,]$ is a $3 \times 4$ matrix known as the canonical projection matrix. The extrinsic camera matrix, $\mathbf{T}_{cw}$ is an element of the Special Euclidean group $SE(3)$ and describes the change of basis from the world coordinate frame into the camera frame. The canonical projection matrix projects the point onto a normalized image plane lying at the unit focal length.

Finally, we have the intrinsic camera matrix, $\mathbf{K}$, which describes the transformation of the normalized image plane into the image plane which is what we typically work with, namely, in pixels units with the origin in the top left corner. It consists of the following parameters, 
\begin{equation*}
    \mathbf{K} = 
    \begin{bmatrix}
    f_x & s   & c_x \\
    0   & f_y & c_y \\
    0   & 0   & 1
    \end{bmatrix} \, .
\end{equation*}

The parameters $f_x$ and $f_y$ are typically referred to as focal lengths and are expressed in pixels per meter and can hence be thought of as a description of pixel density in the sensor's $x$  and $y$ direction. For most cameras, these are equal, but due to various reasons they might differ. For instance, a sensor with non-square pixels, the use of an anamorphic lens, or if the image plane is not parallel with the focal plane. The latter is sometimes artistically intentional and is referred to as tilt photography.

The $s$ parameter refers to the skew of the image plane, in other words, if the image plane has a rhomboid shape. This is typically zero in modern cameras. 

The parameters $c_x$ and $c_y$ describe the origin of the image plane, also known as the principal point. This point translates the image frame origin to the upper left corner, which is typical when working with digital images.

Additionally, real camera systems have lenses that introduce various distortion. Lens distortions are typically centered at the axis of projection and hence introduced before applying the extrinsic matrix. We express this as the application of the function $\Delta$ to the normalized image coordinate
$\bm{\tilde p}_i$ as  
\begin{equation*}
    \bm{p}_i = \mathbf{K}\Delta(\bm{\tilde p}_i) \, .
\end{equation*}
The distortion function is composed of a set of different types of distortions~\cite{lens-distortion-models}. The most common ones that are considered by default in common camera calibration libraries are radial and tangential distortions. 

Lenses work by refracting light, and the refraction index of a medium is wavelength-dependent, which can be realized by shining white light through a prism. This leads to wavelength-specific lens distortions such as chromatic aberration and chromatic focal shift. Lenses are typically designed with this in mind in order to minimize the effect. Lens distortion corrections used in this paper will be considered independent of wavelength.

\subsection{Camera calibration}
Camera calibration is the task of finding the parameters of the intrinsic and extrinsic matrix. There are various methods, but the gold standard is to use Zhang's method which only requires a planar calibration target, and hence easy to produce to high accuracy~\cite{zhang2000flexible}.

The calibration target employed in Zhang's method can be any planar shape with uniquely identifying features of known relative position such that point correspondences can be determined between the image plane and the world coordinate frame. The most common calibration target is a checkerboard type target, and automatic feature detection is implemented in various software packages and libraries.

The calibration method involves taking ($\geq 3$) photos of the planar calibration target in various poses inside the field-of-view of the camera which gives enough constraints to determine all the intrinsic and extrinsic parameters. 

\subsection{Parametric representation of lines \& planes}
A line can be described by the set of points
\begin{equation}
    L = \{ \bm{p} = \bm{l_0} + \lambda \bm{v} \mid \lambda \in \mathbb{R} \} \, ,\label{set:line}
\end{equation}
where $\bm l_0$ is a point on the line, and $\bm v$ a vector which describes the direction of the line. A plane can then be described as the span of two intersecting non-parallel lines, which gives us an analogously set
\begin{equation*}
    \pi = \{\bm{p} = \bm{p_0} + \alpha \bm{v_1} + \beta \bm{v_2} \mid \alpha, \beta \in \mathbb{R} \} . 
\end{equation*}
However, it is more common to use the \emph{point-normal} form:
\begin{equation}
    \pi = \{ \bm{p} \mid \bm{n }\cdot (\bm{p} - \bm{p_0}) =  0 \} \, , \label{eqn:point-normal}
\end{equation}
where $\bm n$ is the normal vector perpendicular to the plane and $\bm{p}_0$ an arbitrary point on the plane. Given that $\bm n = (a, b,c)$, $\bm p_0 = (x_0, y_0, z_0)$, and $\bm p = (x, y, z)$, \eqref{eqn:point-normal} can be written as 
\begin{align*}
        (a, b, c) \cdot (x - x_0, y - y_0, z - z_0) &= 0 \\
        ax + by + cz + d &= 0  ,
\end{align*}
where
\begin{equation*}
    d = -(ax_0 + by_0 + cz_0).
\end{equation*}
Hence, a plane can also be described by the four-vector $\bm{\phi} = (a, b, c, d)^T$. 

\subsection{Laser plane calibration} \label{sec:laser-plane-calibration}
\begin{figure}
    \centering
    \includegraphics[width=0.9\linewidth]{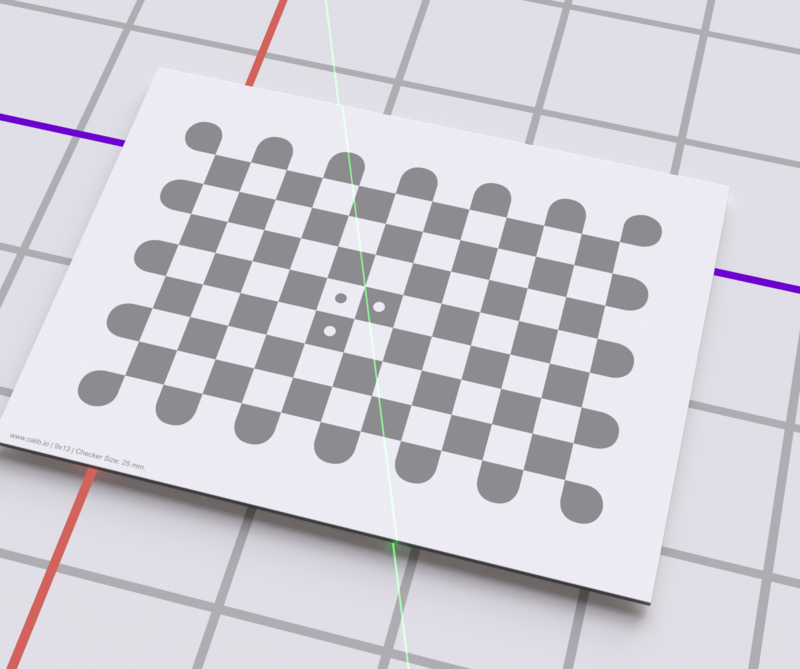}
    \caption{Example of an image used for calibration of the camera and the laser-plane.}
    \label{fig:laser-calibration}
\end{figure}
\FloatBarrier

Similar to how camera calibration is the process of determining the parameters describing the camera, laser plane calibration is the operation of determining the point-normal vector, $\bm{\phi} = (a, b, c, d)^T$. 

The basic principle of laser plane calibration is to identify at least three non-collinear points lying on said plane. For increased precision and robustness to noise, we typically use many more points than this. 

This laser calibration pipeline is similar to that of regular camera calibration. The camera and laser plane calibration can be combined into a single calibration step that utilizes the same images.

Calibration begins by acquiring several pictures of the laser line intersecting the checkerboard target. For each image, we say that the calibration target defines the world coordinate frame of that image and arbitrarily chooses it to lie at $z = 0$. The inner checkerboard corners are then found in the image, for which we then know the corresponding point in world coordinates.

The mapping of the inner checkerboard corners from the calibration plane to the image plane is defined by a planar homography, $\mathbf{H}_{iw}$. Such that
\begin{equation*}
\bm{p}_i = \mathbf{H}_{iw} \bm{p}_w \, ,
\end{equation*}
where $\bm{p}_w = (x_w, y_w, 1)$ and $\bm{p}_i = (x_i, y_i, 1)$ are the world and image coordinates respectively represented in homogeneous form. 
The homography can be estimated using a the direct linear transform (DLT) algorithm~\cite{hartley2004multiple}, which is then further refined by perform the following nonlinear minimization
\begin{equation*}
    \min_{\mathbf{H}_{iw}} \sum_i || \bm{p}_i - \mathbf{H}_{iw}\bm{p}_w ||_2^2 
\end{equation*}
using the Levenberg-Marquardt algorithm~\cite{zhang2000flexible}.

The homography is related to the projection formula 
\eqref{eqn:camera_proj_full} by:
\begin{equation*}
\mathbf{H}_{iw} = \begin{bmatrix}
        \bm{h}_1 & \bm{h}_2 &\bm{h}_3
    \end{bmatrix} 
    = \mathbf{K}
    \begin{bmatrix}
        \bm{r}_1 & \bm{r}_2 & \bm{t}
\end{bmatrix}. \label{eqn:hom-proj-relation}
\end{equation*}
Where the right most matrix is a reduced form of the transformation matrix in \eqref{eqn:camera_proj_full}, where the third column, $\bm{r}_3$, has been discarded. This is valid since we have defined our calibration target to lie at $z = 0$, and any transformation leaves the $z$-component unchanged. Rearranging \eqref{eqn:hom-proj-relation} yields: 
\begin{equation*}
\begin{aligned}
    \bm{r}_1 &= \lambda \mathbf{K}^{-1} \bm{h}_1 \\
    \bm{r}_2 &= \lambda \mathbf{K}^{-1} \bm{h}_2 \\
    \bm{r}_3 &= \bm{r}_1 \times \bm{r}_2 \\
    \bm{t} &= \lambda \mathbf{K}^{-1} \bm{h}_3
\end{aligned}
\end{equation*}
with 
\begin{equation*}
    \lambda = 1/||\mathbf{K}^{-1} \bm{h}_1|| = 1/||\mathbf{K}^{-1} \bm{h}_2||\,.
\end{equation*}
The last column, $\bm{r}_3$, of the rotation matrix can be found as the cross product of the first two columns since elements from $SO(3)$ are orthonormal~\cite{lynch2017modern}. 

Thus we can construct the rotation matrix $\mathbf{\hat R}_{cw}$ as 
\begin{equation*}
\mathbf{\hat R}_{cw} = \begin{bmatrix}
        \, \bm{r}_1 & \bm{r}_2 & \bm{r}_3 \,
    \end{bmatrix}
\end{equation*}

As described by Zhang~\cite{zhang2000flexible}, due to noise, the resulting matrix rarely fulfill the requirements of being a rotation matrix, i.e. $\mathrm{det}\,\mathbf{\hat R}_{cw}^T \neq 1$.
By performing the following minimization, 
\begin{equation*}
    \min_{\mathbf{R}_{cw}} || \mathbf{R}_{cw} - \mathbf{\hat R}_{cw} ||^2_F \, ,
\end{equation*}
with the constraint that $\mathbf{R}_{cw}^T\mathbf{R}_{cw} = \mathbf{I} $. Then we determine the closest proper rotation matrix to $\mathbf{\hat R}_{cw}$, in terms of the squared Frobenius norm. The full transformation $\mathbf{T}_{cw}$ is thus given as
\begin{equation}
    \mathbf{T}_{cw} = 
    \begin{bmatrix}
        \mathbf{R}_{cw} & \bm{t} \\
        \bm{0} & 1
    \end{bmatrix} \, .\label{eqn:transformation-matrix}
\end{equation} 

For each image, the image coordinates of the laser line lying inside the calibration plane is then extracted and refined to sub-pixel accuracy. It can then be projected back onto the the world coordinate plane using the inverse of the previously described homography.
\begin{equation*}
\begin{aligned}
    \bm{p}_w &= \mathbf{H}_{iw}^{-1} \bm{p}_i \\
             &= \mathbf{H}_{wi} \bm{p}_i
\end{aligned}
\end{equation*}
By abuse of notation, $\bm{p}_w$ is subsequently transformed into the camera frame using \eqref{eqn:transformation-matrix}. This is possible since $\bm{p}_w$ has an implicit $z$ coordinate of zero, as was the 
\begin{equation*}
    \bm{p}_c = \mathbf{T}_{cw} \bm{p}_w \, .
\end{equation*}
This is beneficial since the camera frame is a constant frame of reference shared by all views of the calibration plane. 

Once the coordinates of the laser line in each view has been determined and transformed into the camera coordinate frame, we can relatively easily determine the parameters of the plane by setting up the following over-determined homogeneous system of linear equations:
\begin{equation}
\begin{bmatrix}
    x_{11} & y_{11} & z_{11} & 1\\
    \multicolumn{4}{c}{$\vdots$} \\
    x_{ij} & y_{ij} & z_{ij} & 1
\end{bmatrix}
\begin{bmatrix}
    a \\
    b \\
    c \\
    d
\end{bmatrix} = 0 \, ,\label{eqn:laser-eqn-system}
\end{equation}
where $◌_{ij}$ refers to the $j$th coordinate in the $i$th view. Similar to before, and initial guess of the solution can be found using singular value decomposition and then refined further with Levenberg-Marquardt nonlinear optimization. 

We optimize using the following point wise error function $\bm{p}_{ij}$ on the laser line,
\begin{equation*}
    \epsilon_i =  \frac{|\bm{p}_{ij} \cdot \bm{\hat \phi}|}{||\bm{\hat n}||}  + (||\bm{\hat n}|| - 1)^2 \, .
\end{equation*}
Where $\bm{\hat \phi} = (\hat a, \hat b, \hat c, \hat d)^T$ is the estimated plane parameter vector, and $\bm{\hat n} = (\hat a, \hat b, \hat c)$ is the estimated normal vector. 

The first term in the error function is simply the shortest distance from the point to the plane, and the second term is a penalty term that ensures a unit length normal vector to avoid the trivial solution. 

\subsection{3D reconstruction}
Given that we have a calibrated the camera and been able to identify the four parameters that describe the laser plane, then we can fully reconstruct the corresponding 3D position of a point on the image plane which represents the laser line. 

This is possible since any point $\bm{p}_i = (x_i, y_i, 1) \in \mathbb{P}^2$ in the image plane describes a line that intersects the camera origin and the image plane. If $\bm{p}_i$ also corresponds to a point in the image plane where the laser line is visible, then the line it describes must also intersect with the laser plane at some point, $\bm{p}_c$. This point is what we want to recover since it lies on the 3D geometry we wish to know the position of.

In order to express the point $\bm{p}_i$ as a line in space, we first need to transform it into the normalized image plane. This is achieved by multiplying it with the inverse camera matrix,  $\mathbf{K}^{-1}$. 
\begin{equation*}
    \bm{\tilde p}_i = 
    \begin{bmatrix}
        \tilde x_i \\
        \tilde y_i \\
        1
    \end{bmatrix}
    = \mathbf{K}^{-1}
    \begin{bmatrix}
        x_i \\
        y_i \\
        1
    \end{bmatrix}
\end{equation*}
where
\begin{equation*}
    \mathbf{K}^{-1} = 
    \begin{bmatrix}
        \frac{1}{f_{x}} & - \frac{s}{f_{x} f_{y}} & \frac{c_{y} s}{f_{x} f_{y}} - \frac{c_{x}}{f_{x}}  \\
        0 & \frac{1}{f_{y}} & - \frac{c_{y}}{f_{y}} \\
        0 & 0 & 1
    \end{bmatrix}
\end{equation*}

The normalized image coordinate, $\bm{\tilde p}_i$, represents the direction vector, $\bm{v}$, in the line equation \eqref{set:line}. And since this line intersects the origin, we select that as our arbitrary point $\bm{l}_0$. This means that the intersection, $\bm{p}_c$, between this line and the laser plane must be a scalar multiple of 
\begin{equation}
    \bm{p}_c = \lambda \bm{v} = \lambda \bm{\tilde p}_i \label{eqn:point_in_camera}
\end{equation}
Substituting this into \eqref{eqn:point-normal} gives us
\begin{equation*}
    \bm{n} \cdot ( \bm{l_0} + \lambda \bm{v} - \bm{p_0} ) = 0 \, .
\end{equation*}
And if we solve for $\lambda$ we get: 
\begin{equation}
    \lambda = \frac{\bm{p_0} \cdot \bm{n}}{\bm{v} \cdot \bm{n}} \label{eqn:lambda}
\end{equation}
(Note that if $\bm{v} \cdot \bm{n} = 0$, then the plane and the line is parallel to each other, and if $\bm{l_0} = \bm{p_0}$, then the line is contained inside the plane. These cases can easily be avoided in a real-world system, by making sure that the laser plane is: not parallel with the camera's $z$-axis, and not intersecting the camera origin.) 

The point of line-plane intersection can then be calculated by inserting~\eqref{eqn:lambda} into~\eqref{eqn:point_in_camera}, which yields:
\begin{equation}
    \bm{p}_c = \frac{\bm{p_0} \cdot \bm{n}}{\bm{v} \cdot \bm{n}} \, \bm{\tilde p}_i \, . \label{eqn:reconstruction_formula}
\end{equation}
If the plane is known only only in its four vector form, $\bm{\phi} = (a,b,c,d)^T$, then $\bm{p}_0$ can be calculated as 
\begin{equation*}
    \bm{p}_0 = -\frac{d\bm{n}}{||\bm{n}||^2} \, , 
\end{equation*}
which represents the point closest to the plane.

\section{Implementation} \label{sec:implementation}
The laser scanner itself is implemented inside of Blender version 2.92, but older versions are also supported. We utilize both the built-in render engine Cycles, as well as the external render engine, LuxCoreRender, which is made available inside of Blender via the BlendLuxCore plug-in. Some features were also using the Python API which is accessible via the built-in scripting interface. Software related to calibration, 3D reconstruction, etc. were all implemented as Python scripts outside of Blender. 

\subsection{Virtual Camera}
The virtual camera is Blender's default camera type with its parameters set to match a physical one. The emulated camera is an Omron Sentech STC-MCS500U3V equipped with a 12 mm lens set to an aperture of $f/1.4$ and focused at 1 meter. The camera outputs images with a resolution of 2448$\times$2048, which with a physical pixel size of \SI{3.45}{\micro\meter} corresponding to a sensor size of 8.4456$\times$7.0656 \si{\milli\meter}.

\begin{figure*}[ht!]
    \centering
    \begin{subfigure}[t]{0.24\linewidth}
    \includegraphics[width=0.95\linewidth]{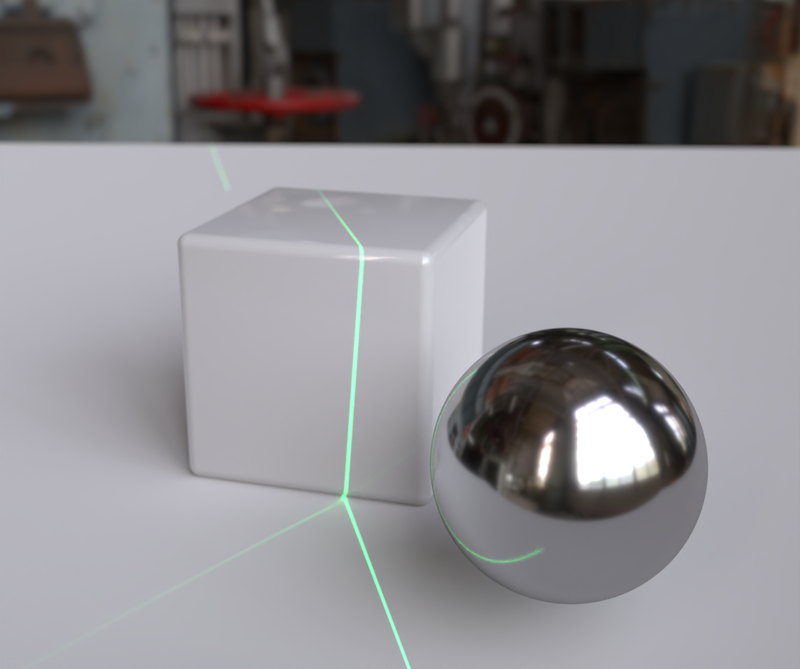}
    \caption{RGB}
    \label{fig:output-data:rgb}
    \end{subfigure}
    \begin{subfigure}[t]{0.24\linewidth}
    \includegraphics[width=0.95\linewidth]{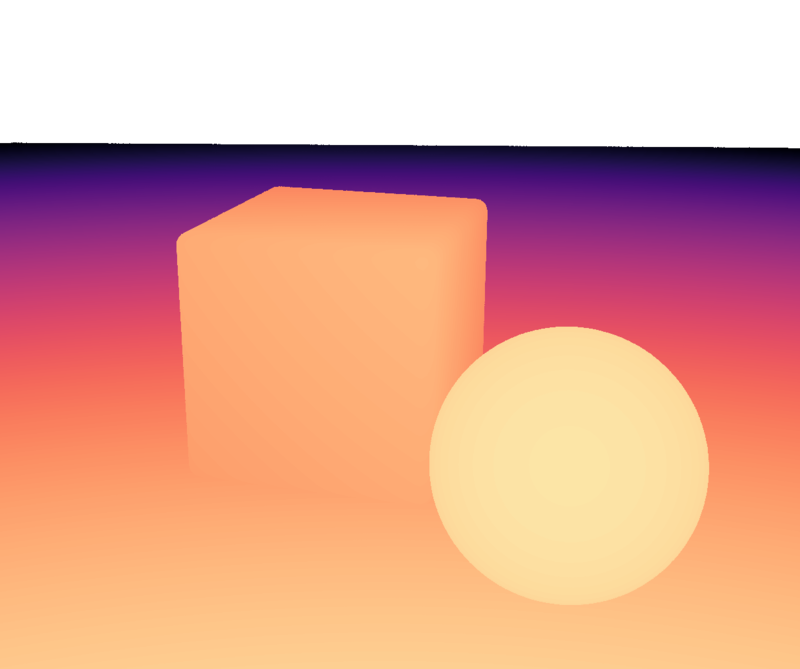}
    \caption{Depth}
    \label{fig:output-data:depth}
    \end{subfigure}
    \begin{subfigure}[t]{0.24\linewidth}
    \includegraphics[width=0.95\linewidth]{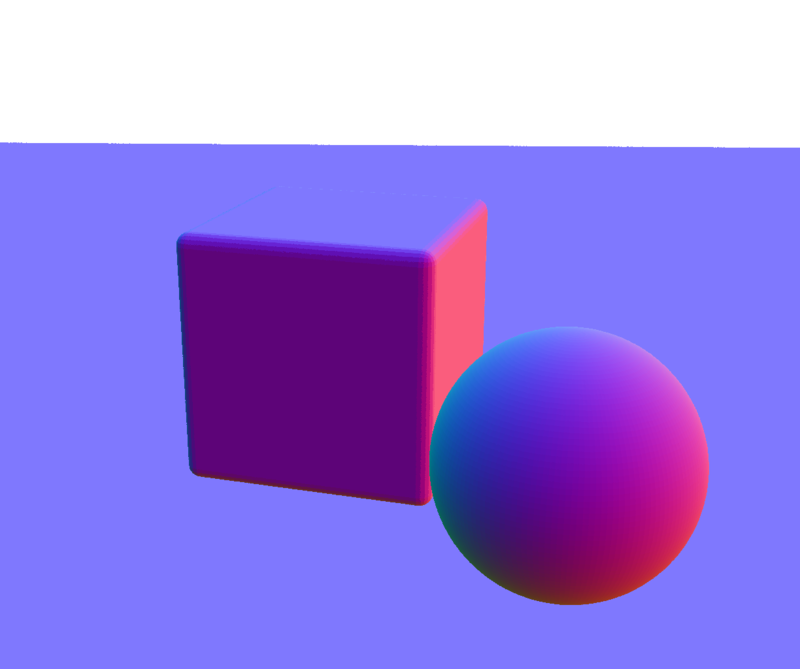} 
    \caption{Surface normals}
    \label{fig:output-data:normal}
    \end{subfigure}
    \begin{subfigure}[t]{0.24\linewidth}
    \includegraphics[width=0.95\linewidth]{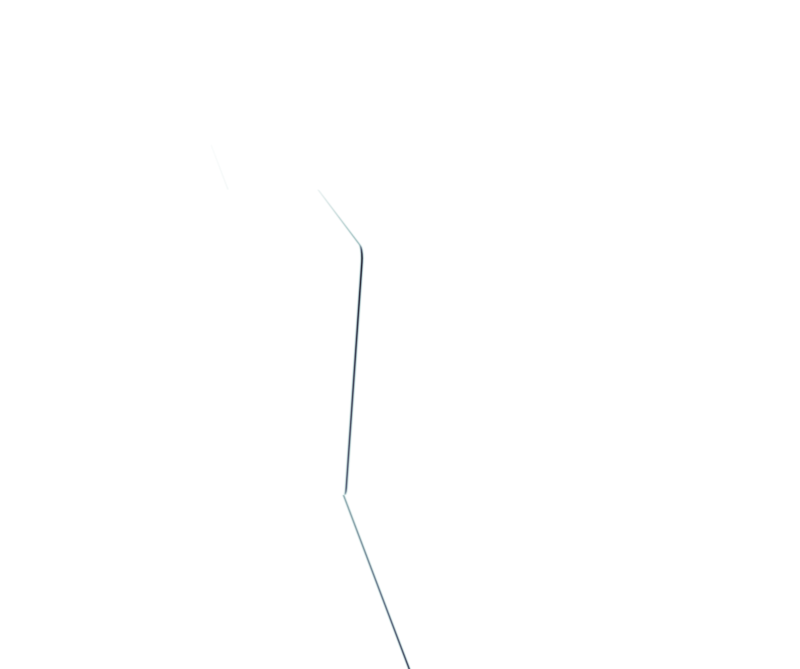}
    \caption{Ground truth laser mask}
    \label{fig:output-data:lasermask}
    \end{subfigure}
    \caption{Example of the output data produced.}
    \label{fig:output-data}
\end{figure*}

\subsection{Virtual laser}
The virtual laser is implemented as a generic line laser, albeit inspired by a model Z20M18H3-F-450-lp45 line laser from Z-LASER GmbH, which has an output power of 20 mW at 450 nm. Neither Cycles~\cite{blender} nor LuxCoreRender~\cite{luxcorerender} is a spectral rendering engine, and hence the specific wavelength is not directly applicable. Instead, the color of the virtual laser is specified by eye. Due to differences in functionality between the two render engines, the line laser was implemented in two different fashions.

\subsubsection{Cycles}
In the Cycles render engine, the laser line is implemented by using a spotlight with an emission shader and a laser intensity mask. The mask is generated procedurally inside Blender by building up a node network, which is a form of visual programming. The procedural approach makes the virtual laser very flexible and can easily be tuned to match a real one.

A spotlight in Blender radiates light spherically from its origin, but only light inside a cone centered around the negative $z$-axis is emitted onto the scene. The direction of the light can be accessed via the normal vector, which we first normalize by dividing by the $z$-component. This essentially projects the vector onto the $z=1$ plane.

The intensity profile of the cross-section of a line laser is typically Gaussian. By applying a Gaussian function to our normal vectors w.r.t. the $x$-component results in an intensity mask that represents our laser line. In this case, we use a unit amplitude Gaussian which has the form: 
\begin{equation*}
    g(x) = \exp\left(- \frac{x^2}{2 \sigma^2} \right) \, ,
\end{equation*}
where $\sigma$ determines the perpendicular spread of the laser line.

The resulting mask, however, results in a physically inaccurate power value. To make the virtual laser perceptually similar to a real one, the power needs to be specified in watts, rather than milliwatts. In the spirit of physical correctness, a correction factor is calculated such that the value has a physical meaning.

The cause for this is due to how a spotlight is implemented in Blender. The power specified does not represent the visually emitted power, but rather the output power of a point light source. This might be artistically useful, as changing the cone angle does not change the light intensity, but not in this case. To compensate for this we must scale the intensity mask such that the intensity integral is equivalent to that of the unit sphere. 

The point light source has an intensity of one in all directions, and its integrated intensity is the surface area of a unit sphere, i.e $4\pi$. 

For the laser line we find the integral as follows
\begin{equation*}
    \int_{-\gamma}^{\gamma}\int_{-\infty}^{\infty} g(x) \, dx\: dy = 2 \gamma \sigma\sqrt{2\pi} \, . 
\end{equation*}
Where $\gamma = \tan(\theta_{c}/2)$ which is the maximum $y$-value of the projected normal vectors given the specified spotlight cone angle, $\theta_c$. 

In the $x$-direction we the improper integral as an approximation of the proper Gaussian integral which is valid since the length of the laser is much larger than its width and because the Gaussian quickly decays to zero. In the $y$-direction we simply integrate over the length of the laser line.

The intensity mask is then scaled by $\lambda$ which is then the fraction
\begin{equation*}
    \lambda = \frac{4\pi}{2\tan(\theta_{c}/2)\sigma\sqrt{2 \pi} } \, .
\end{equation*}
Hence, a physically correct intensity.

The $\sigma$ value is distance dependant, and instead lasers are commonly specified in terms of their divergence angle, $\theta_l$. The relationship between this angle and the $\sigma$ value is
\begin{equation*}
    \sigma = \frac{\tan(\theta_l/2)}{\sqrt{-2 \ln(1/e^2)}} \, ,
\end{equation*}
which can be used to set the virtual laser to a known value.

\subsubsection{LuxCoreRender}
LuxCoreRender does not support the use of nodes on light sources, and the line laser cannot be procedurally generated. Instead, a picture representation of a line laser is produced in an image editing software coloring the middle column of pixels in the needed color. Gaussian blurring can be applied but does not seem to improve the rendered appearance of the laser line. The image can then be projected using the spotlight directly. This method of creating a virtual line laser is for obvious reasons not as flexible as that of the Cycles engine and might require tuning by eye to get a laser line that matches the physical one.

\subsection{Rendering}
Rendering of the scene is separated into two renderings. The first rendering produces the realistic RGB image that is expected but also contains other passes such as depth, and surface normals. The second rendering disables any lighting in the scene except for the laser and only renders direct reflections. The luminance value of this rendering corresponds to the ground truth location of where the laser intersects the scene. The renderings are then saved in raw form using the multilayer OpenEXR file format. In addition, the RGB image is also saved in PNG for easy viewing since most operating systems come with built-in viewers for this format. Examples of these output images are shown in \cref{fig:output-data}. For each render, a Python Pickle file is also generated, which contains the camera matrix and the laser plane parameters. Other data can easily be extracted from Blender via its Python API and included if needed.

\section{Method} \label{sec:method}

\begin{figure*}[ht!]
    \centering
    \begin{subfigure}[t]{0.32\linewidth}
    \includegraphics[width=0.95\linewidth]{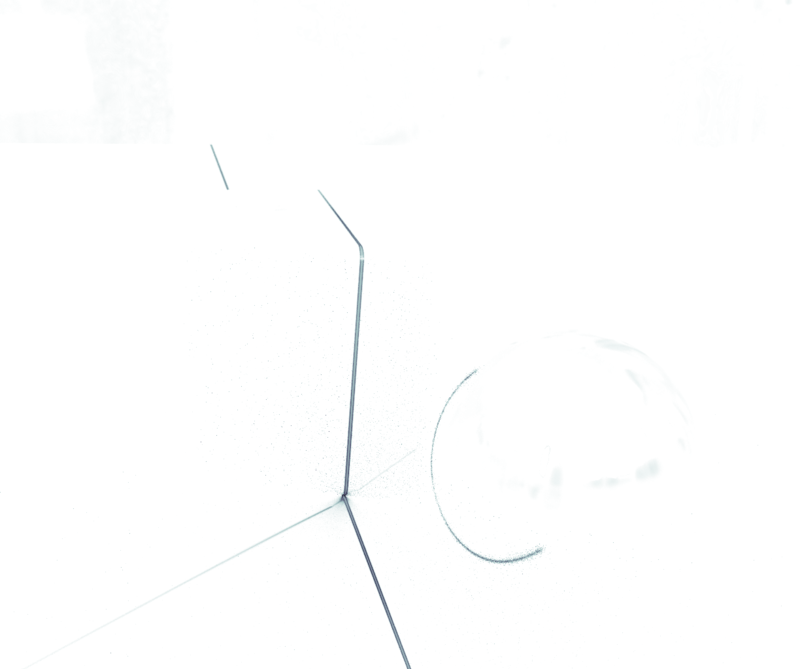}
    \caption{Channel difference of \cref{fig:output-data:rgb} clipped to $I > 0$}
    \label{fig:discrete-pipeline:taubin}
    \end{subfigure}
    \begin{subfigure}[t]{0.32\linewidth}
    \includegraphics[width=0.95\linewidth]{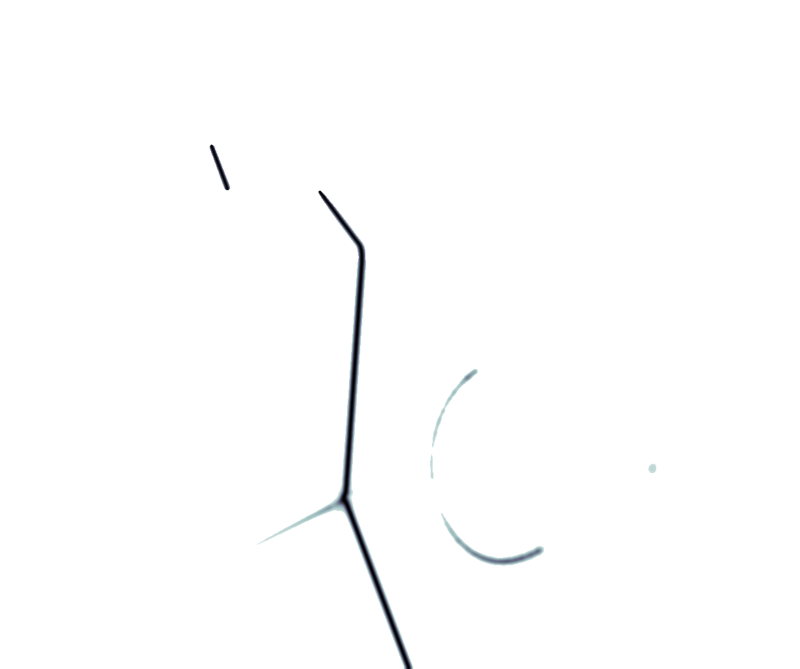} 
    \caption{\cref{fig:discrete-pipeline:taubin} smoothed, thresholded, and row-wise normalized}
    \label{fig:discrete-pipeline:smoothnormal}
    \end{subfigure}
    \begin{subfigure}[t]{0.32\linewidth}
    \includegraphics[width=0.95\linewidth]{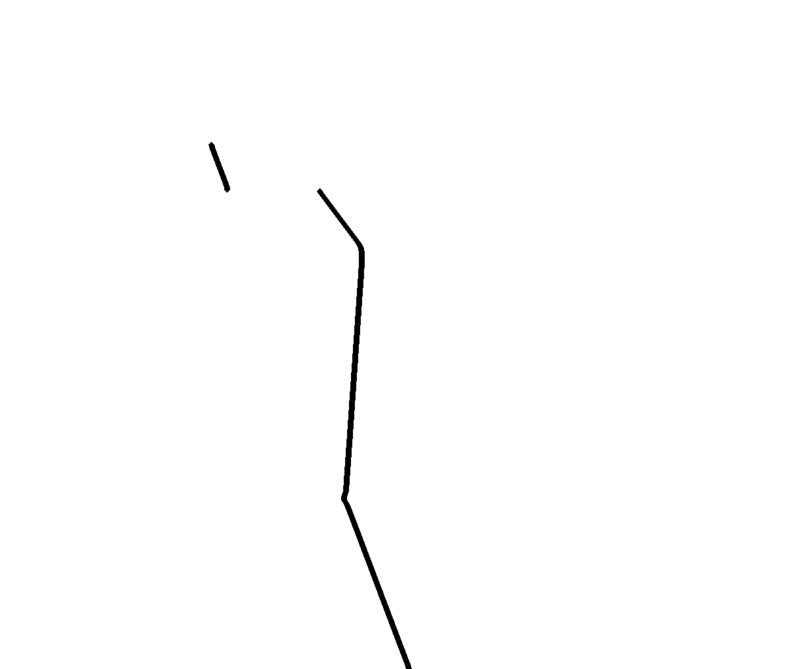}
    \caption{Binary mask of the row-wise maximum of \cref{fig:discrete-pipeline:smoothnormal}. The mask has been dilated for better visibility.}
    \label{fig:discrete-pipeline:rowmax}
    \end{subfigure}
    \caption{Figures illustrating the stages of determining the discrete location of the laser line in the image.}
    \label{fig:discrete_pipeline}
\end{figure*}

\subsection{Camera calibration}
A $13\times9$ checkerboard pattern with 25 mm wide squares and a sheet size of $400\times300$ mm was generated using \href{http://calib.io}{Calib.io} (See \cref{fig:laser-calibration}). The central marker and rounded external checkers were added using Inkscape, an open-source vector drawing software. This central marker allows for partial visibility of the checkerboard and improves overall detection by the \verb|findChessboardCornersSB| function offered by OpenCV. Also note that the saturation of the checkerboard has been reduced, the reasons for this is because the same target is also used for the calibration of the laser plane. By reducing the saturation, absorption of the laser line is avoided. The resulting image is then used to texture a plane of the same size inside Blender to act as a virtual calibration board. With the plane centered on the origin, images of the checkerboard were then rendered with the camera positioned in randomly generated poses above the target.

For calibration of the physical camera, the aforementioned checkerboard pattern was printed using a regular office laser printer and adhered to a sheet of cardboard. In a similar fashion to the virtual camera, several images were acquired from various camera poses. 

Calibration was then performed in Python using the OpenCV library. Feature correspondence was first found using \verb|findChessboardCornersSB()|, and further refined with sub-pixel accuracy. For the physical camera, OpenCV's default camera distortion parameters where used, whereas, for the virtual camera, the lens distortion coefficients were forced to zero since it does not have any lens elements. 
\subsection{Laser plane calibration}
Calibration of the laser plane was only performed in silico. Images of the same virtual calibration target were captured in a similar way as for the camera calibration, albeit ensuring that the laser line was intersecting the target. The plane parameters, $\bm{\phi}$, were then found by using a Python implementation of the process described in \cref{sec:laser-line-extraction} with the aid of the libraries OpenCV, SciPy, and NumPy.

\subsection{Laser line extraction} \label{sec:laser-line-extraction}
The method for extracting the image coordinates of the laser line is a two-step process. First, we find the discrete pixel location of the laser, followed by a sub-pixel refinement by locally fitting a Gaussian function. This is valid as lasers typically have a Gaussian intensity profile.

The discrete pixel location is found by calculating the channel difference image~\cite{taubin20143d}, which for a green laser is defined as
\begin{equation*}
    I_{d} = I_g - \frac{I_r + I_g}{2},
\end{equation*}
but can naturally be arranged to work for red and blue lasers. This amplifies pixels that are primarily green. Values below zero are subsequently discarded before convolving the one-channel image with a $3\sigma$ Gaussian smoothing filter. From the smoothed image, we subtract by the mean intensity and clamp negative values to zero followed by a global normalization to unit intensity. Intensity values of less than 0.1 are subsequently discarded. The discrete laser location is then found by taking the maximum value of the row-wise normalized image. Excerpts from the pipeline can be seen in \cref{fig:discrete_pipeline}.

Sub-pixel accuracy was then achieved by row-wise fitting a Gaussian function to the smoothed channel difference image (before thresholding and normalizing) by using the discrete laser location as an initial location. 

\subsection{Modeling}
For testing of weld profile measurements, a model of a single bevel tee weld joint was modeled inside of Blender. Freely available textures from \url{cc0textures.com} were used to give it a realistic appearance. A ground surface texture was applied in and in the proximity of the weld seam, while the rest was textured to appear slightly rusted.

\section{Results and discussion} \label{sec:results_and_disc}
\subsection{Camera calibration}
The physical camera was calibrated using 30 images resulted in the following camera matrix with a RMS reprojection error of 0.986 px:
\begin{equation*}
    \begin{bmatrix}
        3479.8  & 0         & 1202.3 \\ 
        0       & 3479.0    & 1010.2 \\
        0       &           & 1 
    \end{bmatrix}
\end{equation*}
And with the following lens distortion parameters:
\begin{align*}
    (k_1, k_2, k_3) &= (-0.0785  
    , -0.1658  
    , 2.154) \\ 
    (p_1, p_2) &= (-0.0013 
    , -0.0022) 
\end{align*}

\begin{figure*}[hb!]
    \centering
    \includegraphics[width=0.9\linewidth]{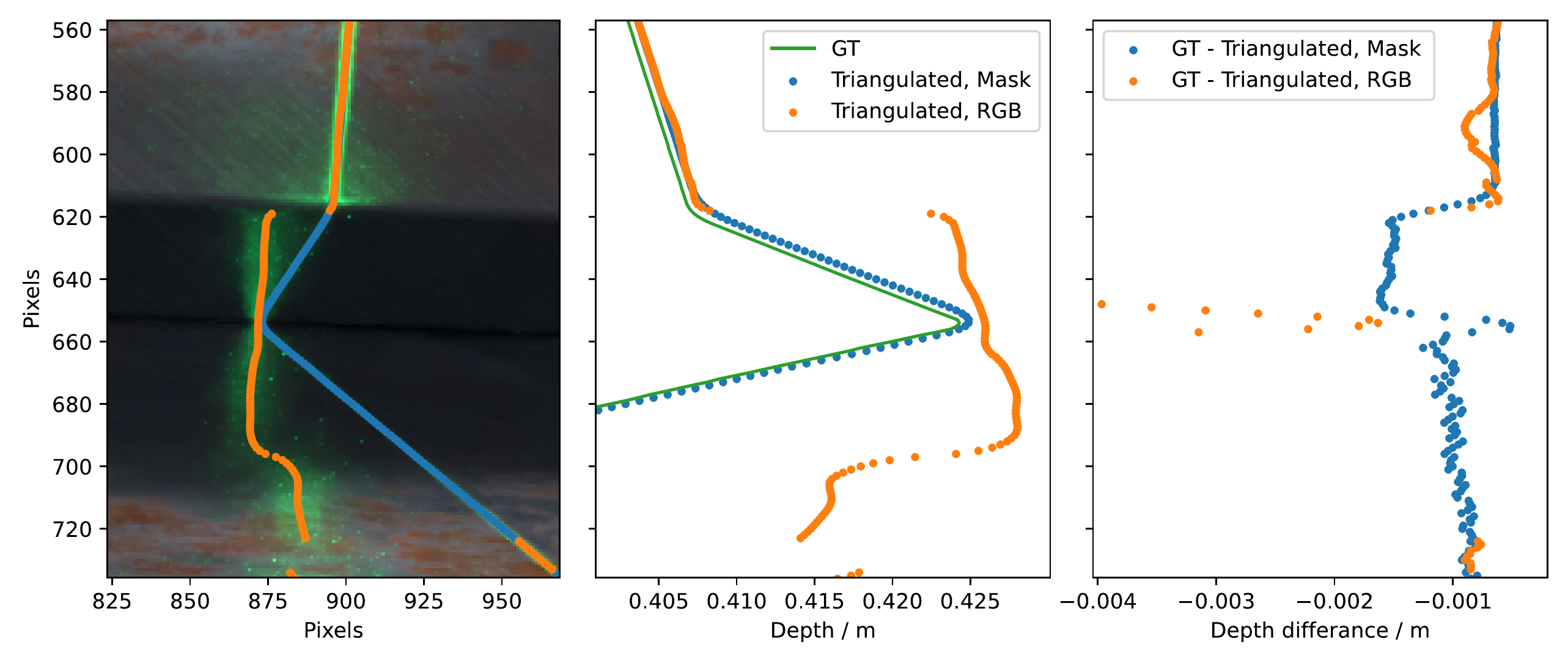}
    \caption{From left to right: 1. The laser line coordinates extracted using either the RGB image, or the laser mask. 2. The $z$-distance relative to the camera. 3. The difference between the distance found by the two triangulation methods and that of the ground truth.}
    \label{fig:weld-seem-recon}
\end{figure*}
The virtual camera, which was calibrated with 38 images using the same procedure, albeit without lens distortion parameters, achieved a reprojection error of 0.058 px, with the camera matrix:
\begin{equation*}
    \begin{bmatrix}
        3478.4  & 0     & 1223.8 \\ 
        0       & 3477.1 & 1021.4 \\
        0       & 0     & 1 
    \end{bmatrix} \, 
\end{equation*}
The true camera matrix from Blender is the following:
\begin{equation*}
    \begin{bmatrix}
        3478.3 & 0      & 1224 \\
        0      & 3478.3 & 1024 \\
        0      & 0      & 1 
    \end{bmatrix}
\end{equation*}

The resulting camera matrix of both the virtual and physical camera are close to the theoretical one. In the physical calibration, we can see a larger discrepancy in the principal point of around 22 and 14 pixels in $x$- and $y$-direction respectively. This indicates that the sensor is not perfectly aligned with the lens's axis of projection. Given that the pixel size, this equates to about 76 and 48 \si{\micro\meter}, which are reasonable manufacturing tolerances. The similarity of the results signifies that the virtual camera is a close virtual representation of the physical one.

\subsection{Laser plane calibration}
For calibration, the virtual laser was placed 20 cm in the $+x$-axis direction of the camera frame with a 13$^{\circ}$ rotation inwards around $y$-axis. This results in a ground truth point-normal vector with the following (rounded) values:  
\begin{align*}
    \bm{\phi}_{gt} = (9.744\times10^{-1},\, &-6.706\times10^{-8},  \\
        &2.249\times10^{-1},\, -1.949\times10^{-1})^T
\end{align*}
Note that the $y$-component should be zero in theory, and the discrepancy most likely comes from rounding errors during floating-point arithmetic.
The estimated (rounded) values after calibration were found as
\begin{align*}
    \bm{\phi}_{est} = (9.743\times10^{-1},\, &4.095\times10^{-4}, \\
        &2.254\times10^{-1},\, -1.954\times10^{-1})^T
\end{align*}
This corresponds to an angle difference of 0.63 mrad (36 millidegrees) between the ground truth and the estimated normal vector. Such a discrepancy corresponds to about a 1 mm difference at one meter from the pivot point. Higher accuracy can most likely be achieved by using utilizing even more calibration images. The results nevertheless illustrate that this method is viable, even when relatively few images are used. 

\subsection{3D reconstruction}
\begin{figure*}[!ht]
    \centering
    \begin{subfigure}[t]{0.32\linewidth}
    \includegraphics[width=0.95\linewidth]{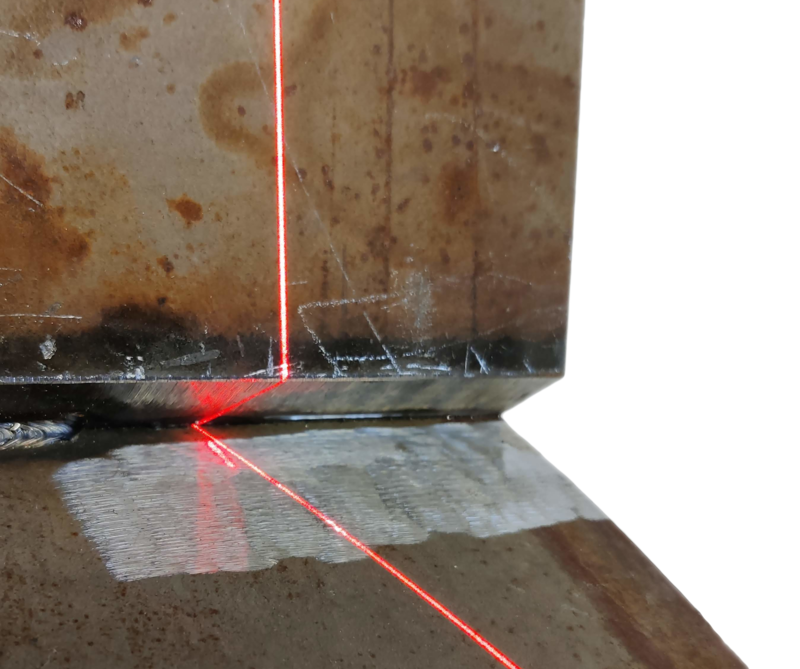} 
    \caption{Real weld seem}
    \label{fig:weld-seem-caustics:real-weld-seem}
    \end{subfigure}
    \begin{subfigure}[t]{0.32\linewidth}
    \includegraphics[width=0.95\linewidth]{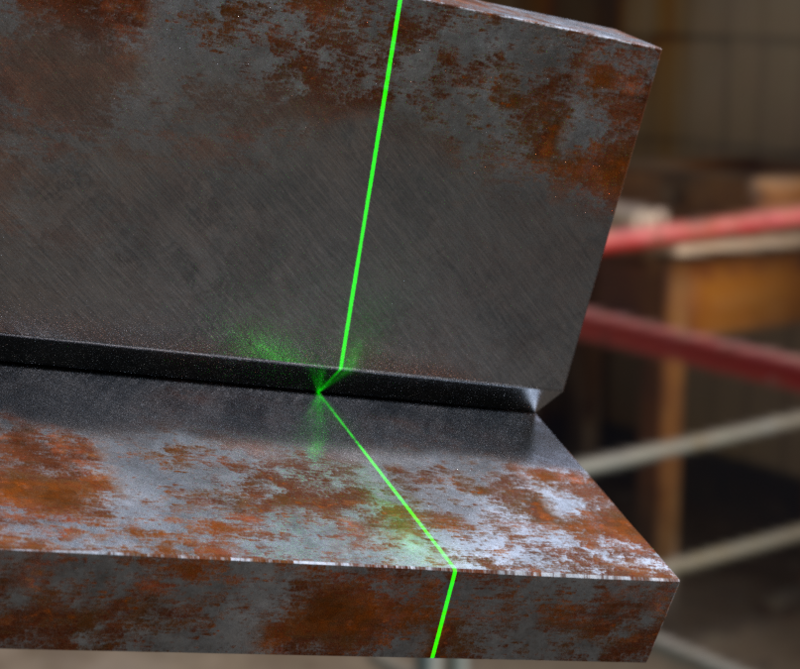}
    \caption{Virtual weld seem}
    \label{fig:weld-seem-caustics:virtual-weld-seem1}
    \end{subfigure}
    \begin{subfigure}[t]{0.32\linewidth}
    \includegraphics[width=0.95\linewidth]{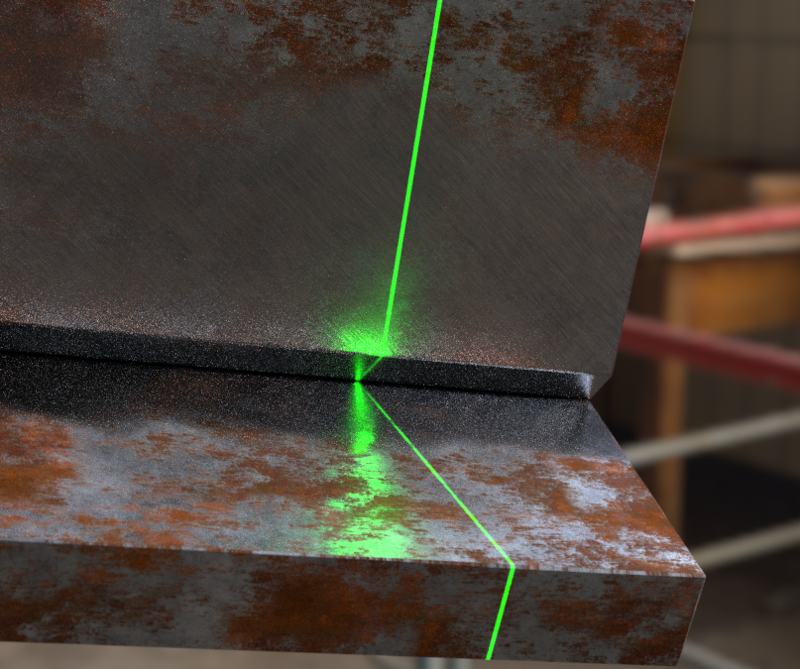}
    \caption{Virtual weld seem from a slightly different angle}
    \label{fig:weld-seem-caustics:virtual-weld-seem2}
    \end{subfigure}
    \caption{Comparison of the reflections when shining a line laser on a real single bevel tee weld joint versus our simulated samples. The simulated scenes were rendered using the LuxCoreRender engine which handles caustics better than Cycles.}
    \label{fig:weld-seem-caustics}
\end{figure*}

As an example, a profile of a weld seam similar to that of  \cref{fig:weld-seem-caustics:virtual-weld-seem1} was scanned virtually. The depth is found in three different ways: By triangulating the depth using the coordinate extracted from the RGB image. Through triangulation by using the laser mask (e.g., \cref{fig:output-data:lasermask}), or by using the ground truth depth image. For triangulation, the ground truth plane parameters were used. 

The ground truth was calculated by using the sub-pixel accurate locations of the laser mask. The depth values at these coordinates were recovered by interpolating the depth output image. In this example, we used a distance-weighted average of a local $3\times3$ window as the interpolation function. 

As can be seen in the leftmost plot in \cref{fig:weld-seem-recon}, the naive laser line extraction method being used is not adequate for reflective surfaces. From the second two plots, we can also see that there is a systematic error occurring since the distance error is always negative even without the presence of severe reflective distortion (e.g. top of the image). This bias could stem from the laser extraction method but is not evident upon inspection of the extracted coordinates. This indicates that there might be a bias in either the ground true depth image, or the plane parameters. The plane parameters could for instance be wrong if the virtual laser line projection is off-center from the $z$-axis. The average (normalized) difference vector between the points triangulated using the ground truth laser mask (blue dots in the plot), and the points reconstructed from the depth image was $(-0.19,  0.01,  0.98)$. This suggests that an off-center projection is indeed the culprit, since the $x$-component is significantly large. Further investigation is nevertheless required. 

During the process of making these figures, it was also discovered that the output from by the $z$-pass given LuxCoreRender does not refer to the distance in $z$ direction from the camera, but rather the Euclidean distance from the camera. LuxCoreRender does however offer a `position'-pass that returns the 3D coordinate for each pixel. The correct depth can easily be recovered from that and is what is used as ground truth in the figure.

In this study, we have not compared the results of the virtual system with that of a real system since this would require highly accurate and calibrated targets. Such a method is described in VDI/VDE 2634(2), which is a standard for evaluating structured light 3D scanners~\cite{3d-precision}. The standard could be used to further validate this system.

\begin{figure}[ht]
    \centering
    \includegraphics[width=0.8\linewidth]{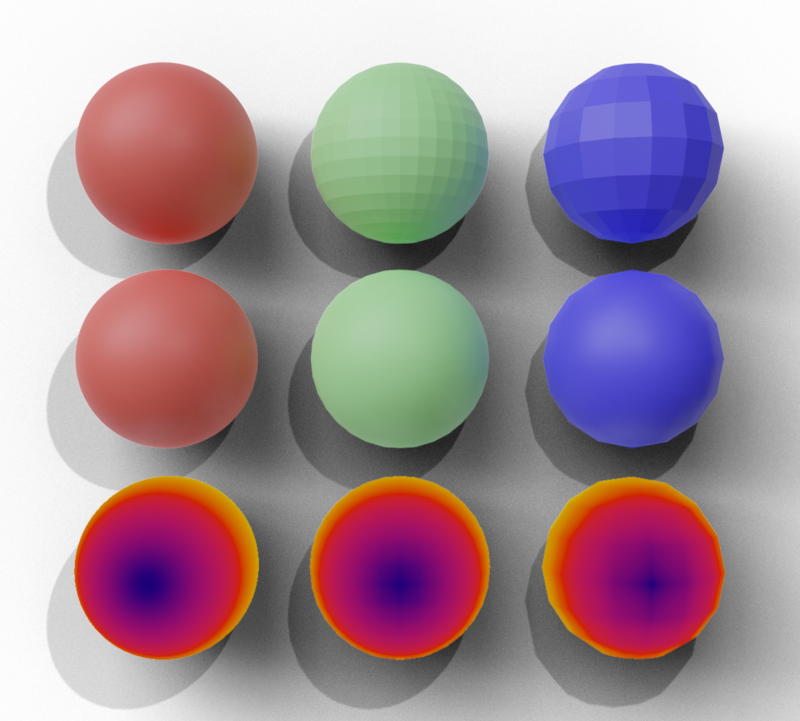}
    \caption{UV spheres of various vertex count where each column has the same number of vertices. They are then rendered as: flat (top), smooth (middle), and in the bottom row their corresponding depth map is visualized.}
    \label{fig:subdivisions}
\end{figure}

\subsection{Reflections}
One of the potential applications of this system is for simulated laser scanning of weld seems. On physical weld seems, rust and other surface contaminants are often ground away before welding which yields a relatively reflective surface. As mentioned in the introduction, reflected laser light is difficult to filter out. For this system to be applicable for research in this area we need to ensure that these reflections are captured well. \cref{fig:weld-seem-caustics} shows a comparison between a real weld seem sample and renderings of a virtual weld seem, illustrating that the system does indeed handle these reflections properly when the right render engine is used.

\begin{figure}[hb]
    \centering
    \begin{subfigure}[t]{0.49\linewidth}
    \includegraphics[width=0.95\linewidth]{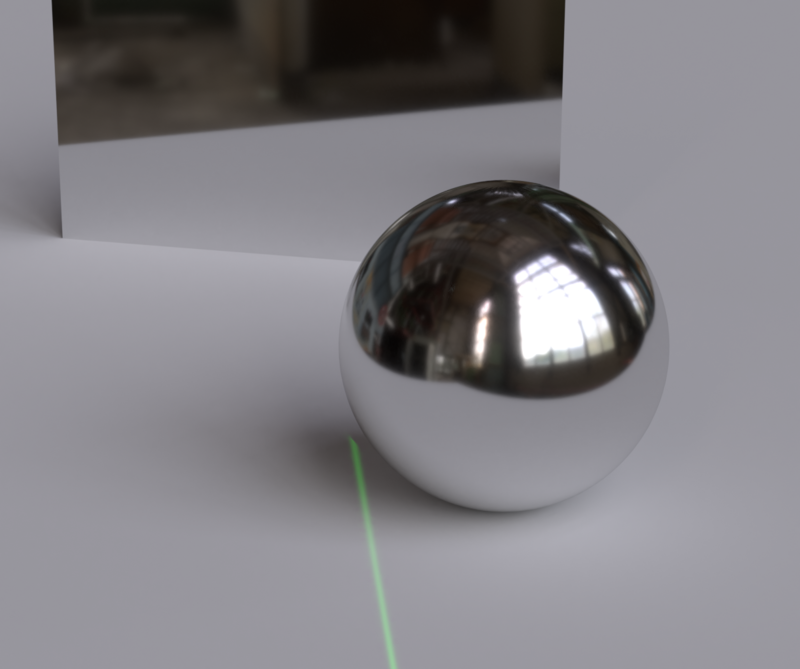} 
    \caption{Cycles}
    \label{fig:cycles-caustic}
    \end{subfigure}
    \begin{subfigure}[t]{0.49\linewidth}
    \includegraphics[width=0.95\linewidth]{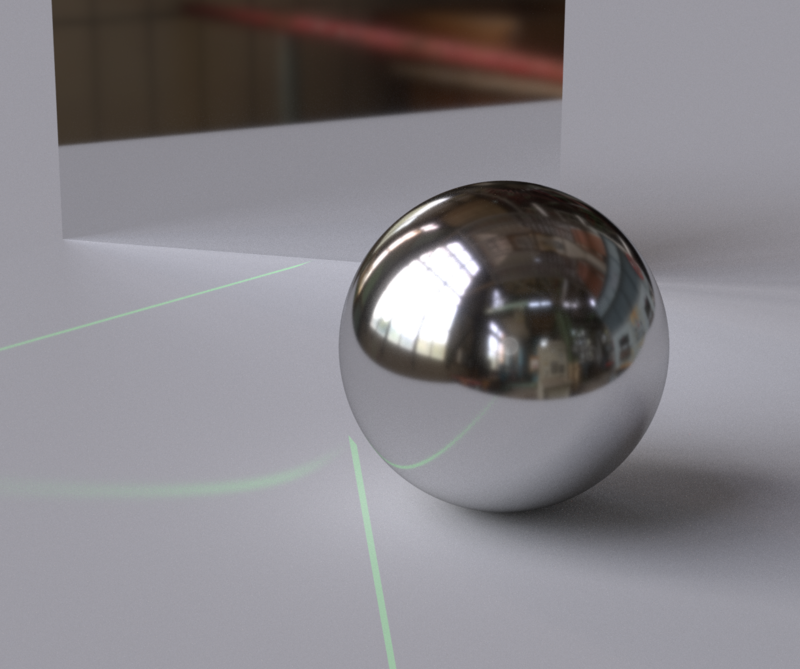}
    \caption{LuxCoreRender}
    \label{fig:luxcore-caustic}
    \end{subfigure}
    \caption{Example scene that visualizes how the two different render engines handle caustic reflections. Both scenes were rendered with 512 samples.}
    \label{fig:lux-vs-cycles-caustic}
\end{figure}

In computer graphics, reflections of this type are referred to as `caustics'. Path-tracing rendering algorithms in general do not handle caustics that well due to the stochastic nature of the algorithm. It is particularly bad when small light sources are used, such as in the case of a line laser, since the probability of a light ray reflecting into the laser is very low. LuxCoreRender, compared to Cycles, performs much better at this because it enables bidirectional path-tracing when a sufficiently glossy material is encountered. I.e. light rays are also traced from light sources to that material. As can be seen in \cref{fig:lux-vs-cycles-caustic}, LuxCoreRender produces realistic-looking reflections whereas Cycles completely fails at capturing these. 

\subsection{Laser mask}
The ground truth laser position is currently produced as a mask which must be further processed to give a sub-pixel accurate laser line location. This might be adequate, but further work should be done to evaluate this. A better option is to find a method for extracting the sub-pixel location directly from Blender which would remove any form of ambiguity. 

\subsection{Lens distortion}
The virtual camera inside of Blender is an ideal pinhole camera and hence does not suffer from lens distortion. This is not a physically accurate representation, and since the goal is to produce data with the least amount of reality gap as possible, it might be needed to include this in the simulation. It is currently only possible to add rudimentary lens distortion as a post-processing step, which might not be a good solution since this would introduce interpolation artifacts. However, when working with a real system, you typically correct your images to remove distortion which also an interpolation step. 

\verb+pbrt+~\cite{pbrt} is another render engine that has the ability to model a lens as a stack of optical elements. The dimensions of the optical elements are typically not disclosed by the manufacturer, so it is unclear how useful this is.

Another option is to use differentiable rendering techniques, such as those described in \cite{mitsuba2}, to generate a virtual lens that produces the same distortion as that of the real lens.

A final approach would be to disregard the traditional camera model and use a generalized camera model which uses a per-pixel parametrization of the camera~\cite{10000params}. This would completely remove the problem of having to consider the complexities of lens distortions. This would be possible to implement in Blender since it is open-source, but would nevertheless require substantial efforts.

\subsection{User limitations}
A significant limitation of this implementation lies at the user end. In order to produce a virtual object that matches a real one requires both knowledge about 3D modeling, as well as some artistic skill and expertise when choosing the correct texture and BSDF parameters. 

This could possibly be mitigated in the future as differentiable techniques mature. For instance, in \cite{shi2020match} and \cite{reparameterizing}, they utilize such techniques to reconstruct geometry, texture, and material properties from a sample.

A specific example of where a naive approach might yield poor results is shown in \cref{fig:subdivisions}. The outward appearance of the middle row is quite similar, despite that the underlying geometry is vastly different in the number of faces. In the second row, the shading is set to `smooth' where the normal of a certain point on the geometry is an interpolation of the neighboring vertex normals which results in a smooth surface appearance. This has the benefit of achieving a visually appealing render while keeping the number of faces to a minimum and hence reducing the render time. However, as can be seen in the final row where the depth values are visualized, the choice of shading does not change the resulting depth and hence the produce ground truth depth would not be accurate. 

Regardless of these user limitations, the papers presented in \cref{sec:related_works} clearly show that even simple synthetic data can be used for machine learning applications. Instead of thinking of synthesized data as being the sole source for training, it might be better to think of it as a form of augmentation to be used in conjunction with real data.

\section{Conclusion} \label{sec:conclusion}
In this paper, we have presented Blazer, a virtual 3D laser scanner implemented in Blender\textsuperscript{\textregistered}. It leverages both built-in and external physically based rendering engines to create labeled data for training a neural network or validating traditional methods. The implementation is made available at Github under the MIT license to enable other researchers to benefit from this implementation.

The system consists of a virtual camera and line laser, which are implemented to match a real-world system closely. In one of the line laser implementations, care was taken to ensure that the specified parameters had a physical meaning.

Through experiments, we show that a virtual camera can be simulated to match a real camera's parameters to a high degree. Lens distortion was not included, but we presented suggestions on how this can be implemented and accounted for. The laser plane was also calibrated using traditional methods resulting in parameters with a 36 millidegree plane normal inaccuracy to the ground truth, which indicates that the rendered images are an accurate representation of the virtual system parameters. 

The limitations of the system and their potential impact on its suggested applications were discussed, such as how the choice of render engine can result in non-physical results. Ideas on how differentiable rendering techniques could mitigate some of these limitations were also presented and a method for further validation of the system is proposed.

\section{Future work} \label{sec:future}

The current implementation of the pipeline requires much knowledge about the intricacies of Blender and the implementation itself. To increase the usability, the authors aim to make this implementation a part of the BlenderProc framework, which would help remove these intricacies via abstraction layers. This will make this work more accessible to researchers working in the sim2real field who might already be familiar with the BlenderProc API. 

Future work also entails using synthetic data to train neural networks targeted towards laser scanning tasks. Early tests on laser line extraction trained only with synthetic data show very promising results at transferring the knowledge to the real domain.

\section*{Acknowledgments}
The first author wants to thank Ola Alstad for all the valuable discussions. Andreas Theiler for his helpful comments on the manuscript. And finally, CG Matter for his inspiring content that made this paper possible. 

{\small
\bibliographystyle{abbrvnat}
\bibliography{paper}
}

\end{document}